\pdfoutput=1
\documentclass{article}

\usepackage[preprint]{neurips_2024}

\usepackage{fancyhdr}
\usepackage{nth}
\usepackage[utf8]{inputenc} 
\usepackage[T1]{fontenc}    
\usepackage{hyperref}       
\usepackage{url}            
\usepackage{booktabs}       
\usepackage{amsfonts}       
\usepackage{nicefrac}       
\usepackage{microtype}      
\usepackage{xcolor}         
\usepackage{graphicx}
\usepackage{hyperref}
\usepackage{amsmath}
\usepackage{cleveref}
\usepackage{color, colortbl}
\usepackage{tabularray}
\usepackage{multirow}
\usepackage{nicematrix}
\usepackage{caption}
\usepackage{subcaption}
\usepackage{booktabs} 
\usepackage{wrapfig}
\usepackage{fancyhdr}
\usepackage{hyperref}
\definecolor{mygray}{gray}{0.9}

\title{Drug Repurposing Using Deep Embedded Clustering and Graph Neural Networks}

\definecolor{mydarkblue}{rgb}{0,0.08,1}
\definecolor{mydarkgreen}{rgb}{0.02,0.6,0.02}
\definecolor{darkred}{rgb}{0.8,0.02,0.02}
\definecolor{darkorange}{rgb}{0.40,0.2,0.02}
\definecolor{darkpurple}{RGB}{111,0,255}
\definecolor{myred}{rgb}{1.0,0.0,0.0}
\definecolor{mygold}{rgb}{0.75,0.6,0.12}
\definecolor{mydarkgray}{rgb}{0.66, 0.66, 0.66}

\author{%
  \parbox{\textwidth}{\centering
  Luke Delzer\textsuperscript{1}, Robert Kroleski\textsuperscript{1}, Ali K. AlShami\textsuperscript{1}, Jugal Kalita\textsuperscript{1} \\
  \textsuperscript{1}University of Colorado Colorado Springs \\
  \texttt{ldelzer@uccs.edu}, \texttt{rkrolesk@uccs.edu}, \texttt{aalshami@uccs.edu}, \texttt{jkalita@uccs.edu}}
}

\begin{document}

\maketitle

\begin{abstract}
Drug repurposing has historically been an economically infeasible process for identifying novel uses for abandoned drugs. Modern machine learning has enabled the identification of complex biochemical intricacies in candidate drugs; however, many studies rely on simplified datasets with known drug-disease similarities. We propose a machine learning pipeline that uses unsupervised deep embedded clustering, combined with supervised graph neural network link prediction to identify new drug-disease links from multi-omic data. Unsupervised autoencoder and cluster training reduced the dimensionality of omic data into a compressed latent embedding. A total of 9,022 unique drugs were partitioned into 35 clusters with a mean silhouette score of 0.8550. Graph neural networks achieved strong statistical performance, with a prediction accuracy of 0.901, receiver operating characteristic area under the curve of 0.960, and F1-Score of 0.901. A ranked list comprised of 477 per-cluster link probabilities exceeding 99 percent was generated. This study could provide new drug-disease link prospects across unrelated disease domains, while advancing the understanding of machine learning in drug repurposing studies.
\end{abstract}

\section{Introduction}
\label{sec:introduction}
Drug discovery and development is notoriously expensive and time-consuming. Traditional pharmaceutical pipelines often span more than a decade and require upwards of \$2 billion to bring a single new drug to market \cite{Baig:16, Schlander:21}. Consequently, approximately 90\% of candidate compounds fail to reach regulatory approval \cite{waring2015analysis}, leaving behind a trove of under-exploited data on abandoned drug candidates. Even approved drugs are rarely re-evaluated for novel therapeutic uses beyond initial indications. These challenges have fueled growing interest in drug repurposing, the process of identifying new disease indications for existing or previously abandoned drugs as an alternative to de novo development.

Recent advances in Machine Learning (ML) have opened new avenues to accelerate drug repurposing \cite{Vamathevan:19, Yang:22, ghandikota2024application,alshami2024smart}. Modern ML methods, especially deep learning, can analyze large-scale biomedical datasets to uncover complex patterns linking drugs and diseases. A particularly promising direction is to leverage multi-omics data, an integrative collection of diverse biological data modalities (e.g., genomics, transcriptomics, proteomics, metabolomics, etc.), that together provide a more comprehensive view of molecular and clinical information \cite{Tanoli:21, Mohammadzadeh-Vardin2024}. By integrating multi-omics profiles, ML models can identify subtle biologic signatures and mechanistic commonalities that suggest repurposing opportunities (e.g. a shared pathway or target between an existing drug and a different disease).

However, effectively harnessing multi-omics data for systematic drug repurposing presents several core challenges. Data fragmentation and heterogeneity are major issues. Although an abundance of chemical and omics data exist, resources are often siloed by study or disease area and lack a unified structure for integration \cite{ghandikota2024application}. Public databases typically focus on specific experiments or single therapeutic domains, resulting in disconnected datasets that do not readily merge into a comprehensive multi-omic profile for each drug. Moreover, data are frequently organized by broad categories (such as organ systems), which complicates the linking of drug properties to clinical outcomes across disease domains.

Additionally, the high dimensionality of multi-omics features presents challenges for computational models. A single drug can be described by thousands of molecular and phenotypic features, making learning algorithms prone to overfitting. Many prior studies mitigate this by reducing feature complexity or restricting analysis to known biomarkers and pathways \cite{Vamathevan:19}. While such simplifications make analysis tractable, they risk overlooking unexpected associations and limit the discovery of truly novel drug–disease relationships. As a result, existing computational approaches often struggle to identify new uses for drugs beyond well-characterized disease contexts.

To address these challenges, we propose a novel hybrid unsupervised–supervised machine learning framework that combines Deep Embedded Clustering (DEC) with Graph Neural Networks (GNNs) in a unified pipeline. The high-level idea is to utilize DEC to learn compact feature representations from multi-omics data, and then leverage a GNN to predict new drug–disease links based on these learned representations. This approach bridges unsupervised representation learning with supervised graph-based inference.

\begin{figure*}[t]
\centering
\includegraphics[width=\textwidth]{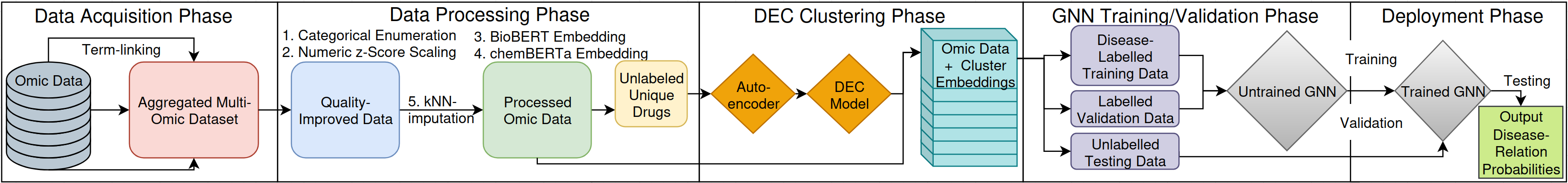}
\caption{Flowchart of study approach: DEC clustering and GNN link prediction are divided into a two-stage pipeline. DEC clusters drugs using multi-omics features, producing embeddings used to construct a drug–disease graph. A GNN is trained on this graph to predict novel drug–disease links.}
\label{fig:study_flowchart}
\end{figure*}

In the first stage of the pipeline, we employ DEC to learn a low-dimensional embedding of the integrated multi-omics feature space and to cluster drugs based on their latent representations. DEC utilizes a deep autoencoder network to perform feature compression and cluster assignment simultaneously. Training DEC on the aggregated multi-omic dataset yields a concise latent vector for each drug, automatically grouping drugs with similar profiles. This unsupervised step reduces noise and dimensionality while uncovering latent data structures that enable clustering of similar drugs.

In the second stage, we construct a heterogeneous graph where nodes represent drugs (augmented with DEC-derived features) and diseases, and edges denote known drug–disease treatment links. We then train a GNN to predict the likelihood of new drug–disease connections (i.e., to identify plausible missing edges). The GNN leverages both the network structure (for example, drugs in the same DEC cluster often treat related diseases) and the DEC-based feature embeddings to inform its predictions. By combining learned feature similarities with known interaction patterns, the model can propagate information and generalize across disease domains. For instance, if a drug clusters with known anti-cancer compounds, the GNN may predict an association between that drug and a similar cancer-related disease. Figure~\ref{fig:study_flowchart} demonstrates the study workflow integrating the DEC-GNN pipeline.

This work addresses the issue of prohibitively expensive drug development costs. While computational methods for drug repurposing have improved, challenges remain due to fragmented multi-omic data and limitations in generalizing across drug and disease domains. This study addresses the problem of finding novel drug-disease associations by combining unsupervised feature extraction and clustering with supervised link prediction across heterogeneous multi-omic data to overcome limitations in feature representation and drug-disease diversity.

In this work, we propose a hybrid machine learning pipeline that integrates DEC with GNNs, combining unsupervised feature learning and supervised link prediction for drug repurposing. We unify heterogeneous multi-omics datasets into a single representation for over 9,000 drugs, enabling cross-disease analysis and addressing data fragmentation. DEC is employed to extract salient low-dimensional features from high-dimensional data, enhancing downstream GNN performance. We achieve state-of-the-art results with approximately 90\% accuracy and an AUC of ~0.96, identifying numerous high-confidence drug–disease links as repurposing candidates.
\section{Related Work}
\label{sec:related_work}
Previous studies have applied deep learning to drug repurposing, particularly using multi-omic data. While prior work has employed feature reduction, clustering, or GNNs, few integrate them or focus on inter-disease repurposing. 

A cancer drug repurposing model, DeepDRK, was developed using kernel methods to integrate multi-omic features and a deep neural network to classify drug responses \cite{10.1093/bib/bbab048}. A similar method was performed in \cite{9669314}, where multi-omic integration was followed by GNN classification. Likewise, improvements on DeepDRK were shown in \cite{Mohammadzadeh-Vardin2024}, which used autoencoders for dimensionality reduction and a multilayer perceptron for cancer drug classification. Results demonstrated improvements over standard drug prediction models by enhancing feature extraction before classification.

Unrelated cancer drugs were clustered to identify novel purposes in \cite{Hameed2018}, which used Growing Self-Organizing Maps and graph clustering to successfully reclassify 39 drugs for novel uses. Similarly, hierarchical clustering was used in \cite{Guo:20} to identify shared mechanisms of action based on chemical properties for Traditional Chinese Medicine.

The GDRnet model was introduced in \cite{Doshi2022}, using a multi-layer heterogeneous GNN with link prediction for disease classification. GDRnet outperformed baseline GNN models, including GraphSAGE and Graph Attention Networks, in identifying candidate COVID-19 treatments. Similarly, a deep learning model combining hierarchical and spectral clustering with a GNN was presented in \cite{bansal2023clusteringgraphdeeplearningbased} to identify FDA-approved drugs with potential activity against COVID-19. Using multi-omic properties, 15 existing drugs were identified with similarities to those awaiting approval for COVID-19 treatment.

A GNN-based approach incorporating Multiple Prior Knowledge (MPK-GNN) was introduced in \cite{Xiao:23} for multi-omics data analysis. The pipeline consists of four interconnected modules that sequentially incorporate prior knowledge to extract deeper latent features, while optimizing both contrastive and task-specific loss. This framework supports a range of downstream multi-omics analysis tasks and demonstrated significant improvement in the Cancer Genome Atlas dataset.

\section{Datasets}
\label{sec:dataset}

Features for 17,616 drugs were aggregated from multiple multi-omic subdomains using publicly available data from five major sources, including CTD\cite{ctd}, PubChem\cite{pubchem}, BioSNAP~\cite{biosnap}, OpenFDA~\cite{openfda}, and the Human Metabolome Database (HMDB)\cite{hmdb}. A summary of these data sources is provided in Table \ref{dataset_summary}. All data was aggregated into a monolithic set of drugs and associated features. Drug features were combined into single instances linked by medical subject headings, ChemicalID number, and omic domain-specific links. Details of the sub-datasets, extracted features, and labels linking drug features across datasets are provided in Appendix~A (see supplementary material). The supplementary materials are available online.\footnote{\url{https://github.com/LDELZ/icmla2025-paper-458-drug-repurposing}}

Data was first acquired from the Comparative Toxicogenomic Database (CTD) due to its comprehensive properties across numerous omic subdomains of interest, including genomic, metabolic, and transcriptomic data. Chemo-omic, protein-omic, and metabol-omic data were appended to each unique CTD drug instance using the PubChem, BioSNAP, and Human Metabolome Databases (HMDB). FDA disease indications were appended from the OpenFDA database to be used as disease labels, and CTD drug-disease links were preserved for unapproved drugs. Duplicate features from among differing datasets were reduced to single attributes and combined cause-effect features were divided into individual attributes.

Drug data quality was determined by calculating the percent of available omic data per instance. Instances falling below 70\% feature completeness were purged from the dataset to reduce noise, minimize imputation, and balance completeness with preservation.  Missing data from among the preserved drug instances was imputed using k-Nearest Neighbors (kNN). KNN was selected for its ability to preserve data relationships among complex multi-omic properties, therefore minimizing unrealistic similarities that could compromise clinical validity. 

\begin{table}[t]
\centering
\resizebox{\linewidth}{!}{%
\begin{tabular}{|p{5cm}|c|c|p{6cm}|}
\hline
\textbf{Database} & \textbf{\# Drugs} & \textbf{\# Features} & \textbf{Multi-omic Contributions} \\
\hline
Comparative Toxicogenomics Database (CTD, \cite{ctd}) & 15,065 & 50--100 & Gene/protein interactions, expression data, toxicogenomics \\
\hline
PubChem Database \cite{pubchem} & 119,000,000\textsuperscript{a} & 500 & Chemical/molecular fingerprints, bioactivity \\
\hline
BioSNAP Network Dataset \cite{biosnap} & 4,510 & 25--50 & Drug-protein and disease-gene associations \\
\hline
OpenFDA Drug Records \cite{openfda} & 17,449 & 100 & Real-world drug use, off-label insights \\
\hline
Human Metabolome Database (HMDB, \cite{hmdb}) & 114,100 & 130 & Drug metabolism, metabolite interactions, biochemical pathways \\
\hline
\multicolumn{4}{l}{\footnotesize$^{\mathrm{a}}$PubChem entries were limited to CTD-sourced drugs to ensure high feature completeness and exclude infeasible industrial compounds.} \\
\end{tabular}%
}
\caption{Overview of datasets and multi-omics coverage.}
\label{dataset_summary}
\end{table}

Categorical data was enumerated for simplification and numeric data was standardized using z-score normalization. Rich linguistic data consisting of chemical mechanisms of action (MOA), gene MOA, and chemical structure representations were converted to textual embeddings. Bidirectional Encoder Representations from Transformers (BERT) embeddings were generated to preserve important relationships from linguistic context. Chemical and gene MOA were encoded using BioBERT, and chemical structures were encoded using ChemBERTa \cite{lee2020biobert,wang2019smiles}, which are BERT-base models trained on biomedical and chemical corpora, respectively. BERT embeddings were appended as new features to each drug instance.

The final dataset was composed of $8,740,262$ disease links, from among $9,022$ unique drugs, each with $2,336$ features. The data was reduced to form a separate clustering dataset consisting of only unique drug instances, without regard to multiple drug-disease links. Disease labels were temporarily removed to perform unsupervised DEC, focusing exclusively on biochemical attributes, while enabling reinclusion of disease links for future GNN evaluation.
\section{Methodology}
\label{sec:approach}

\subsection{\textbf{Deep Embedded Clustering Model}}
\begin{figure}[htbp]
    \centering
    \includegraphics[width=8.2cm, height=4.3cm]{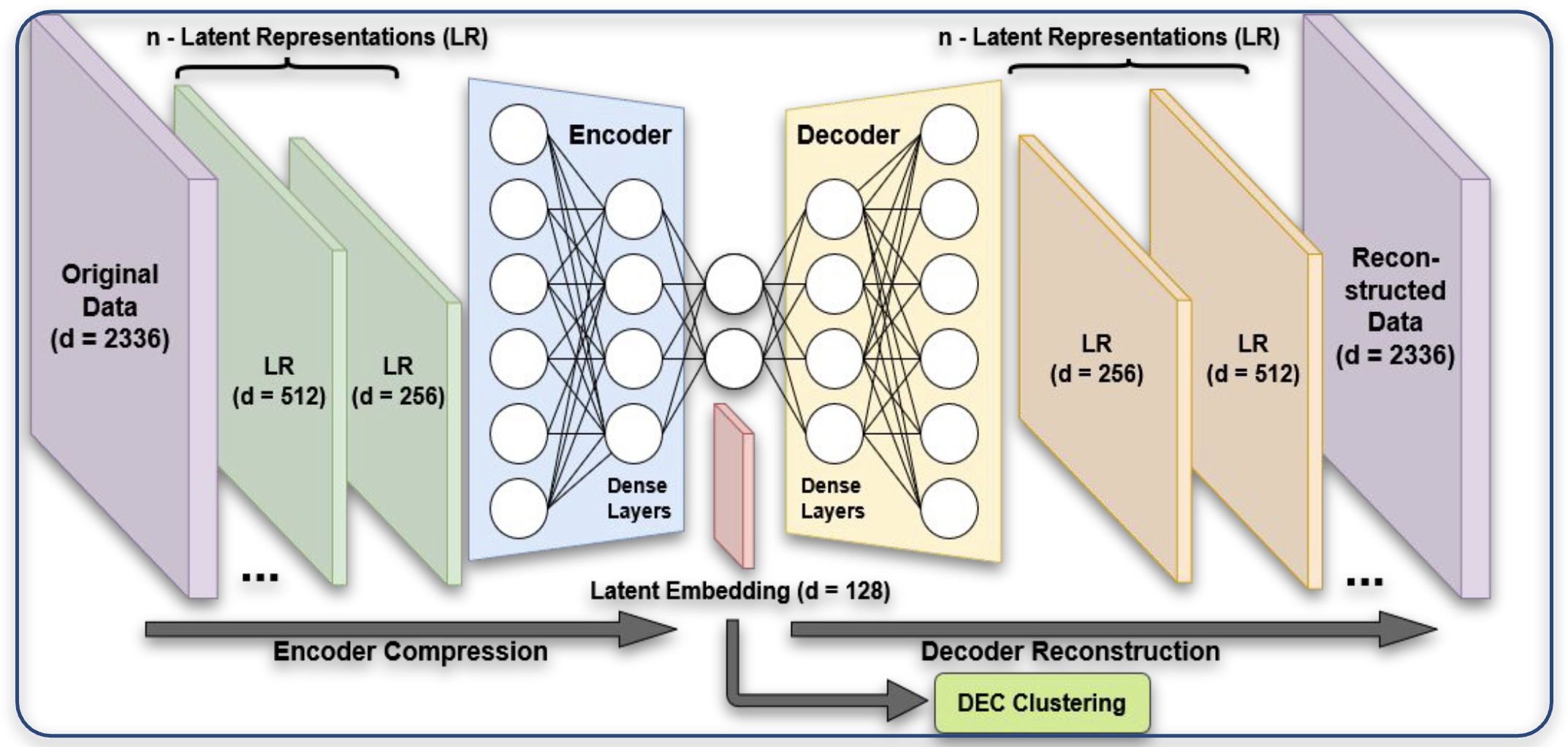}
    \caption{Architecture of the DEC autoencoder. Fully connected layers with ReLU activations perform sequential dimensional compression and reconstruction. The encoder reduces feature dimensionality by half at each layer to form the latent embedding. The decoder reverses this process to calculate loss from the reconstructed data using mean squared error. The optimal latent embedding is preserved for DEC.}
    \label{encode-decode}
\end{figure}

A DEC model was constructed using autoencoders and a symmetric dense layer architecture as displayed in Figure~\ref{encode-decode}. Dense layers of fully connected nodes were used to encode or decode each previous embedding. The Rectified Linear Unit (ReLU) activation function was used in each dense layer to leverage non-linearity and capture intricate combinatorial relationships. Optimization via backpropagation using the Adaptive Moment Estimation (Adam) optimizer supported proper encoding of high-dimensional data \cite{kingma2015adam}. During each encoding phase, a latent embedding (LE) of the previous feature set was generated with half the prior dimensions (d). Sequential dimensional halving enabled preservation of deep omic relationships by controlling the rate of compression until an optimal LE was found. During decoding, the latent embedding doubled the dimensions of each LE until the data was reconstructed to its original size. Reconstruction loss was calculated as the mean squared error (MSE) between the original and reconstructed data.

Model training was conducted over a maximum of 1,000 epochs, with early stopping employed if 10 consecutive epochs demonstrated no loss improvement. Early stopping was used to mitigate overfitting with minimal learning benefit. At convergence, the latent embedding, which is a reduced set of robust drug features, was preserved for DEC analysis.

The trained autoencoder and optimal LE were used to perform DEC. DEC was initialized using k-means clustering, which determines a distribution based on iterative updates to each cluster center and re-assigning of instances based on Euclidean distance. The cluster centers were used with the optimal LE as training data. DEC training was performed by iteratively optimizing the LE and cluster centers based on Student's t-Distribution. From the most confident clusters, a target distribution was calculated, and loss was measured using Kullback-Leibler (KL) divergence between distributions.

The optimized DEC embedding was saved and the instance data was reassociated with the corresponding drug's ChemicalID, along with each instance's cluster identifier. Instance embeddings were reassociated with their known disease links using the ChemicalID, and the reconstructed data was partitioned into subsets containing the instances belonging to each cluster. Cluster subsets were used as separate input data for GNN drug-disease link probability analysis, allowing for specialized groups of homogeneous drugs to be evaluated independently. Clustering helps partition the problem space, reducing noise by grouping biologically similar drugs in latent space, allowing the GNN to focus on homogeneous subsets, and improve link prediction accuracy by modeling each cluster’s relationships separately. This partitioning reduced inter-cluster variance and mitigated noise that could convolute disease link confidence.

The dimensionality of the LE and autoencoder/DEC learning rates were varied in order to optimize reconstruction and KL loss, respectively. MSE and KL loss were plotted against the number of epochs to validate proper learning trends and convergence. DEC cluster quality was optimized by maximizing silhouette score (SS). Values near $+1$ indicate cohesion of instances within clusters, and distinct separation of cluster neighbors.

Configuration changes in the autoencoder were further evaluated by measuring the impacts on the final SS in order to maximize cluster quality across all clustering tasks. Cluster distributions were visualized using t-distributed Stochastic Neighbor Embedding (t-SNE) by projecting multidimensional clusters into 2-dimensions. Distances between instances were plotted by similarity, with intra-cluster points demonstrating the highest similarity, and inter-cluster points lesser similarity.

\begin{figure}[htbp]
    \centering
    \includegraphics[width=0.8\columnwidth]{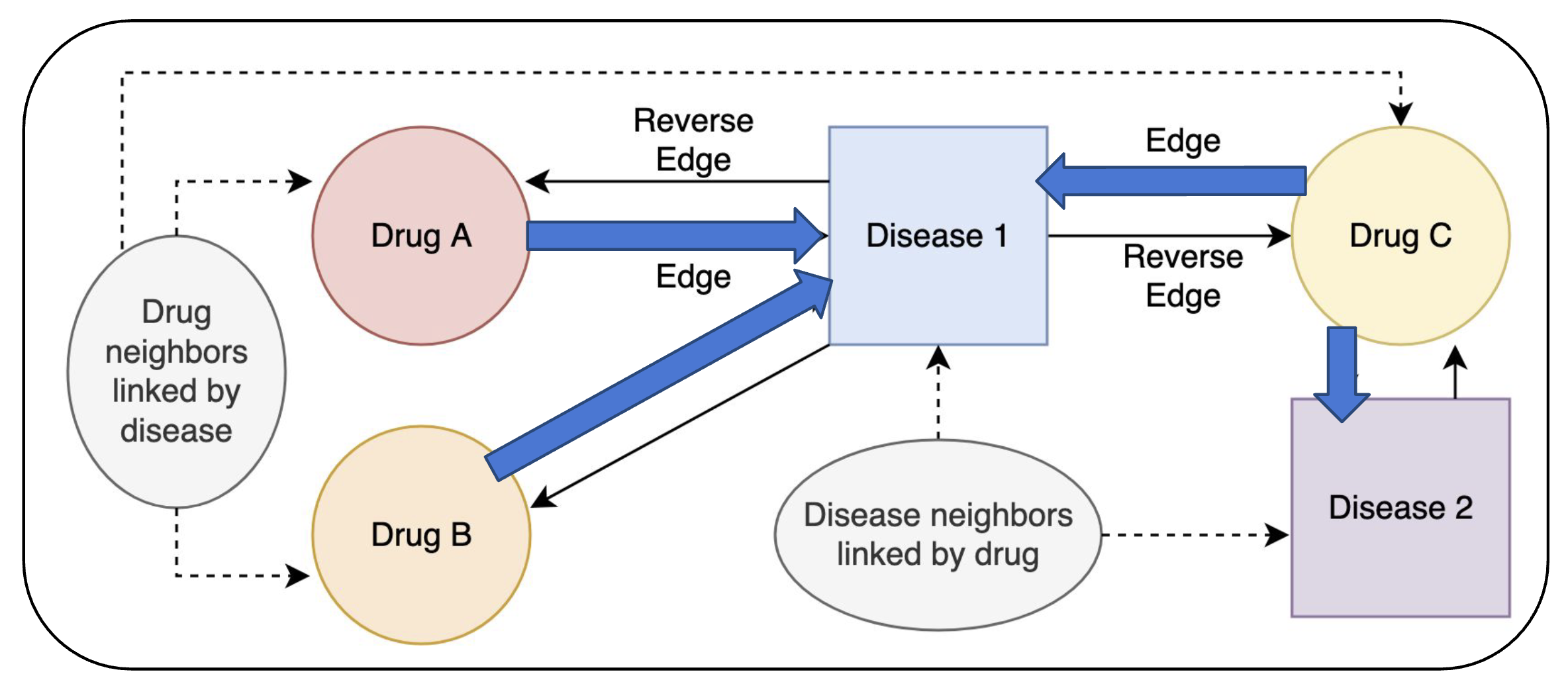}
    \caption{Graph structure used in the GNN model: A bipartite graph of drug and disease nodes with dense edge connections based on known treatment links is shown. Only drug–disease edges are present; drug–drug and disease–disease relations are inferred via 2-hop neighbors.}
    \label{fig:gnn_org}
\end{figure}

\subsection{\textbf{Graph Neural Network Model}}
The drug-disease graph organization is diagrammed in Figure~\ref{fig:gnn_org}. A heterogeneous GNN model was designed by mapping the per-cluster drug instances and disease labels to distinct nodes, forming a bipartite structure. Dense node connectivity was used to account for multiple drugs treating individual diseases and multiple diseases being treatable by individual drugs. Dense connectivity was achieved using direct and reverse edges for each drug-disease relationship. The dataset exclusively provides drug-disease links, and no drug-drug or disease-disease relationships are included. Therefore, each drug and disease form a 1-hop neighborhood to their valid counterpart. Likewise, each drug is related to another drug through a 2-hop neighborhood via a shared disease.

\begin{figure}[htbp]
    \centering
    \includegraphics[width=0.9\columnwidth]{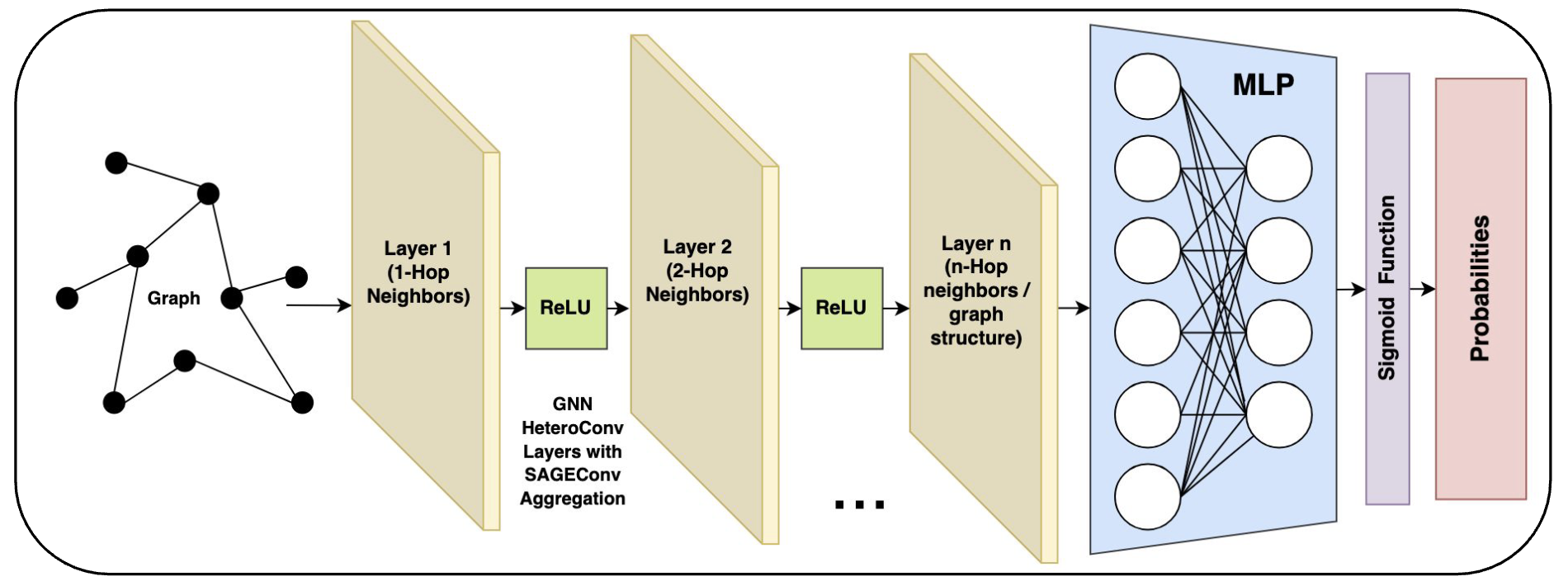}
    \caption{GNN architecture using HeteroConv with SAGEConv layers to propagate and aggregate node information across drug–disease edges. ReLU activations and randomized disease embeddings support nonlinear learning and bidirectional message passing in the bipartite graph.}
    \label{fig:gnn_arch}
\end{figure}

The GNN model architecture is diagrammed in Figure~\ref{fig:gnn_arch}. Multiple convolutional layers were constructed to capture the graph's structural information using HeteroConv \cite{fey2019fastgraphrepresentationlearning}. This enabled message passing of node information between neighbors. Neighbor information was aggregated using sample and aggregate convolutions (SAGEConv) within each convolutional layer, enabling interpretation of structural information \cite{hamilton2018inductiverepresentationlearninglarge}. Aggregation in the first layer creates an embedding for the 1-hop drug-disease neighbors, and additional layers capture broad structural relations from distant neighbors. ReLU activation was used between convolutional layers to introduce nonlinearity for better pattern identification. Since disease labels did not contain feature values, an embedding equal to the drug node dimensionality was stochastically generated with small random numbers. This enabled message passing along the reverse edges to capture drug-drug relationships.

To generate the final link probabilities, an edge decoder was built and applied to the second layer's embedded graph information. For each unique drug-disease link, the aggregated drug and disease node embeddings were stacked and processed by a fully-connected multilayer perceptron (MLP). A dense layer was selected to model complex biochemical relationships between all drug-disease features. Nonlinearity was introduced using the ReLU activation function in the MLP and dropout was applied to improve generalizability through reduced reliance on dominant connections. The sigmoid function was applied to the MLP-generated prediction to determine the probability that each drug independently treats each disease.

Training, validation, and testing datasets were segregated from the optimized DEC embedding in a 70\%-10\%-20\% ratio. The training set represented positive drug-disease associations only, precluding novel disease-link identification. Therefore, an equal number of negative edges were generated from unrelated drugs and diseases. Training was performed iteratively by aggregating neighbor information, calculating the average cross-entropy loss between predictions, and updating model parameters using the Adam optimizer. Early stopping was performed in the same fashion discussed previously. The validation dataset was used for model tuning and the trained-validated model was used to generate the list of novel link probabilities from the test data.

GNN training performance was evaluated by measuring cross entropy loss and predictive performance was evaluated by calculating accuracy, precision, recall, and $F_1$-score. Ranking quality of novel link predictions was assessed by calculating the receiver operating characteristic area under the curve (ROC-AUC), which plots the proportion of true positives against false positives over a range of prediction thresholds. Maximizing ROC-AUC improves the ranking of novel drug-disease associations by ensuring the model correctly distinguishes positive and negative predictions.

A hyperparameter variance analysis was conducted to determine the best GNN parameter configuration that minimizes loss and maximizes statistical performance. Hyperparameters were individually varied and default values were maintained for all other parameters. Default values for tested parameters included LR = 0.001, weight decay = 0, hidden dimensions = 64, GNN layers = 2, and dropout = 0.2. 

ROC-AUC and $F_1$-Score were prioritized when determining the optimal configuration. $F_1$-Score balances precision and recall, which represents that positive predictions are often correct, and true links are identified. This is important because false positives could lead to wasted resources in downstream pharmacologic studies. Additionally, a high number of candidate links is required to ensure that novel uses are found. ROC-AUC best represents the model's ranking ability, which is essential to produce a valid list of drug-disease link probabilities. The optimal configuration was determined using the largest cluster, which maximized training data, statistic stability, and generalizability for identifying novel purposes. This configuration was used to independently train and test new models for each partitioned cluster.

Link probabilities for all drugs in each cluster were calculated using the edge decoder and sigmoid function. Drug-disease links that were already established in the training data were removed in order to isolate novel purposes. The remaining links were ranked by probability, with the highest rankings suggesting the greatest likelihood for novel use. 
\section{Results and Analysis}
\label{sec:experiments}
\subsection{\textbf{Deep Embedded Clustering Results}}

Table~\ref{tab:loss_silhouette_table} demonstrates loss and epoch count at convergence for varied Adam learning rates and latent embedding dimensions used during autoencoder training as well as the associated cluster number and SS at the optimal learning rate. A learning rate of $10^{-4}$ demonstrated superior loss minimization across latent dimensions, with the lowest at 128 dimensions. This LE dimensionality also produced the highest cluster results, with an $SS = 0.8850$ among $35$ clusters.

\begin{table}[htbp]
\centering
\tiny
\renewcommand{\arraystretch}{0.95}
\setlength{\tabcolsep}{3.5pt}
\caption{DEC training and clustering performance across latent dimensions and learning rates.}
\label{tab:loss_silhouette_table}
\begin{tabular}{|c|c|c|c|c|c|c|}
\hline
\textbf{Latent Dim.} & \multicolumn{4}{c|}{\textbf{Learning Rate (Loss / Epoch)}} & {\textbf{k}}&{\textbf{Silhouette\textsuperscript{\tiny a}}
} \\
\cline{2-5}
& $10^{-2}$ & $10^{-3}$ & $10^{-4}$ & $10^{-5}$ & & \\
\hline
8   & 0.9998/13 & 0.9992/11 & 0.9991/11 & 0.9997/11 & 10  & 0.8267 \\
\hline
16  & 0.7758/17 & 0.5067/82 & 0.4818/112 & 0.6647/80 & 6   & 0.7488 \\
\hline
32  & 0.9016/12 & 0.9997/12 & 0.3318/232 & 0.6086/125 & 6   & 0.8019 \\
\hline
64  & 0.6759/29 & 0.2398/128 & 0.1659/348 & 0.4591/169 & 29  & 0.7575 \\
\hline
128 & 0.8552/15 & 0.2760/48 & 0.0999/321 & 0.3148/222 & 35  & \textbf{0.8850} \\
\hline
256 & --          & --          & --           & --           & 12  & 0.7653 \\
\hline
\multicolumn{7}{l}{\footnotesize$^{\mathrm{a}}$ Silhouette score reflects clustering at the optimal learning rate ($10^{-4}$).} \\
\end{tabular}
\end{table}

\begin{figure}[htbp]
    \centering
    \includegraphics[scale=0.34]{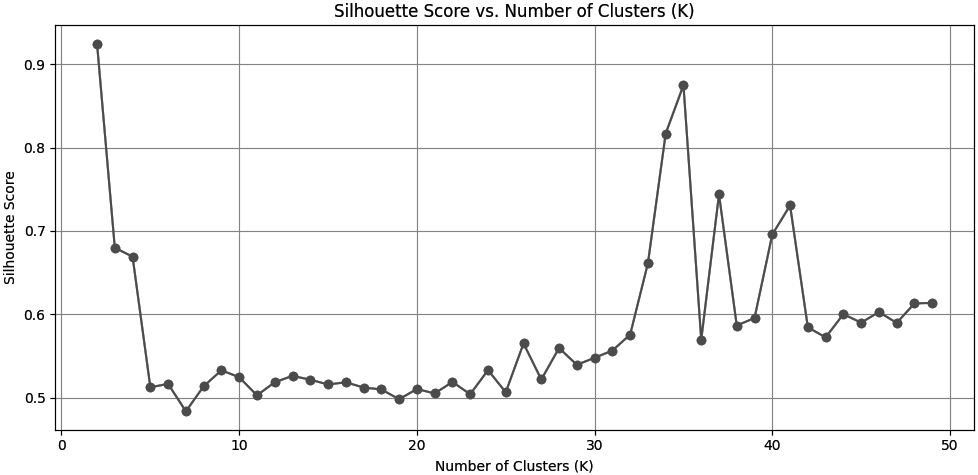}
    \caption{Silhouette scores across cluster counts ($k = 2–50$): A local maximum at $k = 35$ demonstrates a more meaningful distribution for identifying biochemical similarities relevant to drug repurposing.}
    \label{fig:sil_plot}
\end{figure}

Figure~\ref{fig:sil_plot} plots SS against a range of k-values from 2 to 50. While the highest SS was observed with $k = 2$, distributions with excessive overlap result in overgeneralization that precludes repurposing, since dissimilar instances are placed together. A local maximum SS at $k=35$ represents another optimal distribution, which more meaningfully allocates drug instances based on biochemical similarities. Additional local maxima were not explored for higher k-values because significant overlap is required to identify important drug similarities necessary for disease link crossover. Cluster sparseness will limit adequate training data during GNN analysis, potentially reducing repurposing accuracy.

\begin{figure}[htbp]
    \centering
    \includegraphics[scale=0.30]{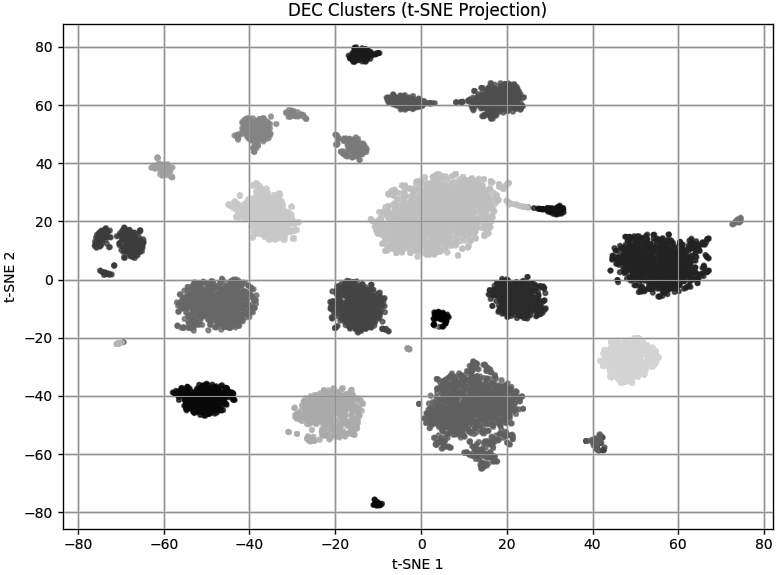}
    \caption{\small{t-SNE visualization of clustered drug instances: Clusters exhibit strong cohesion and clear separation, indicating effective feature compression and high Silhouette Scores, which support their use in downstream GNN analysis.}}
    \label{fig:clusters}
\end{figure}

Figure~\ref{fig:clusters} displays the t-SNE plot of clustered drug instances. The plotted clusters exhibit properties of meaningful clustering, reflecting the robustness of the SS achieved. Instances in each cluster demonstrate cohesion, with tight organization, and sound separation from neighboring clusters. Drug instances within each cluster are the most similar, while those in neighboring clusters are dissimilar. Overlap occurs infrequently, allowing for the partitioning of instances into distinct per-cluster datasets for GNN analysis.

The DEC clustering phase successfully developed a reduced feature embedding representative of the complex, multi-omic data necessary for GNN analysis. The identification of optimized architectural configurations and hyperparameters resulted in a meaningful cluster organization, exhibiting balanced intra-cluster cohesion and inter-cluster separation. The distribution of $9,022$ drug instances into $35$ clusters demonstrates high instance overlap based on biochemical similarities. Each cluster contains a mean of $257.77$ drugs, enabling per-cluster analysis. This distribution implies a high potential for disease link crossover, demonstrating promise for GNN analysis in selecting new diseases from a restricted pool.

\subsection{\textbf{Graph Neural Network Results}}
\begin{figure*}[t]
    \centering
    \includegraphics[scale=0.20]{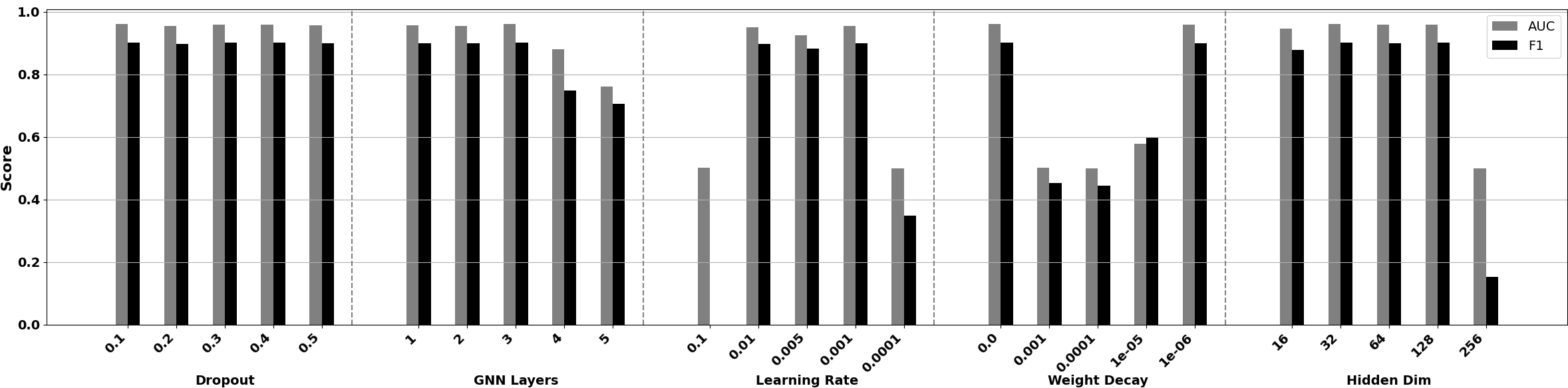}
    \caption{Bar chart of ROC-AUC and F$_1$-Scores across GNN hyperparameter settings, highlighting the impact of dropout, GNN layer depth, learning rate, weight decay, and hidden dimensions on model performance.}
    \label{fig:hp_bar}
\end{figure*}

GNN model optimization was performed using the cluster containing drug instances associated with the highest number of unique drug-disease links. This cluster contained $4,068$ drugs associated with $2,018,434$ disease links, therefore maximizing the amount of training data available for statistical and hyperparameter analyses. Table~\ref{tab:gnn_hp_grid} displays the hyperparameter variance results on prediction performance, and Figure~\ref{fig:hp_bar} plots the ROC-AUC and $F_1$-Scores for each parameter.

The model was notably resistant to changes in dropout, likely due to minimal baseline overfitting. Negative sampling and substantial training data likely made predictions easier to identify, and few GNN layers were required to capture the bipartite structure. This resulted in a relatively simple model that is resistant to overfitting. A low number of GNN layers produced very consistent statistics; however, the best results were seen with three layers. While it is likely that two layers are capable of capturing the bipartite graph structure, the third layer appears to have enabled additional refinement and smoothing of the graph feature representations. Additional layers led to worse performance, likely due to oversmoothing and simplification of graph intricacies. 
\begin{table*}[t]
\centering
\tiny
\caption{GNN hyperparameter tuning results showing performance metrics across varied settings for number of GNN layers, learning rate, weight decay, dropout, and hidden dimensions.}
\label{tab:gnn_hp_grid}
\setlength{\tabcolsep}{5pt}
\renewcommand{\arraystretch}{1}
\begin{tabular}{|c|c|c|c|c||c|c|c|c|c|}
\hline
\textbf{GNN Layers} & \textbf{LR} & \textbf{Weight Decay} & \textbf{Dropout} & \textbf{Hidden Dim} & \textbf{Accuracy} & \textbf{Precision} & \textbf{Recall} & \textbf{$F_1$} & \textbf{ROC-AUC} \\
\hline
1 & 0.001 & 0       & 0.1 & 32  & 0.8974 & 0.8818 & 0.9178 & 0.8994 & 0.9565 \\
2 & 0.001 & 0       & 0.1 & 32  & 0.8961 & 0.8782 & 0.9197 & 0.8985 & 0.9543 \\
4 & 0.001 & 0       & 0.1 & 32  & 0.7993 & 0.7883 & 0.7401 & 0.7474 & 0.8804 \\
5 & 0.001 & 0       & 0.1 & 32  & 0.7331 & 0.7291 & 0.6925 & 0.7063 & 0.7602 \\
3 & 0.01  & 0       & 0.1 & 32  & 0.8932 & 0.8692 & 0.9258 & 0.8966 & 0.9507 \\
3 & 0.005 & 0       & 0.1 & 32  & 0.8744 & 0.8299 & \textbf{0.9417} & 0.8823 & 0.9256 \\
3 & 0.001 & 0.001   & 0.1 & 32  & 0.4985 & 0.4982 & 0.4138 & 0.4521 & 0.5006 \\
3 & 0.001 & 0.0001  & 0.1 & 32  & 0.4994 & 0.4992 & 0.3999 & 0.4441 & 0.5000 \\
3 & 0.001 & 0.00001 & 0.1 & 32  & 0.5832 & 0.5767 & 0.6252 & 0.5999 & 0.5781 \\
3 & 0.001 & 0.000001 & 0.1 & 32 & 0.8988 & 0.8903 & 0.9097 & 0.8999 & 0.9595 \\
3 & 0.001 & 0       & 0.2 & 32  & 0.8955 & 0.8817 & 0.9136 & 0.8974 & 0.9547 \\
3 & 0.001 & 0       & 0.3 & 32  & 0.8988 & 0.8849 & 0.9168 & 0.9006 & 0.9591 \\
3 & 0.001 & 0       & 0.4 & 32  & 0.8987 & 0.8846 & 0.9170 & 0.9005 & 0.9584 \\
3 & 0.001 & 0       & 0.5 & 32  & 0.8968 & 0.8859 & 0.9108 & 0.8982 & 0.9573 \\
3 & 0.001 & 0       & 0.1 & 16  & 0.8764 & 0.8676 & 0.8884 & 0.8779 & 0.9457 \\
3 & 0.001 & 0       & 0.1 & 64  & 0.8981 & 0.8858 & 0.9140 & 0.8997 & 0.9580 \\
\hline
\textbf{3} & \textbf{0.001} & \textbf{0} & \textbf{0.1} & \textbf{128} & \textbf{0.8991} & \textbf{0.8886} & 0.9127 & \textbf{0.9005} & \textbf{0.9605} \\
\hline
\multicolumn{10}{l}{\small $^{\mathrm{a}}$ Bold row highlights the best configuration based on optimal $F_1$ and ROC-AUC scores.}
\end{tabular}
\end{table*}

Learning rate performed poorly with extreme values but consistently with moderate ones. High learning rates likely missed the loss minimum, and low ones failed to converge on a reasonable solution. The intermediate values performed consistently well, likely due to rigorous early stopping. For weight decay, smaller values demonstrated the best statistical performance. This is likely due to minimal noise and high optimization in the DEC latent embedding. Application of high weight decays likely over-suppressed weak associations, leading to worse predictive performance. This observation aligns with how the DEC embedding was demonstrably capable of representing the original feature space during the DEC analysis.  As such, further reduction of the HeteroConv hidden layer dimensionality from the original 128 dimensions demonstrated no meaningful performance decline. Only artificial increasing of the hidden dimensionality led to instability and worse performance. 

\begin{figure*}[t]
    \centering
    \includegraphics[scale=0.29]{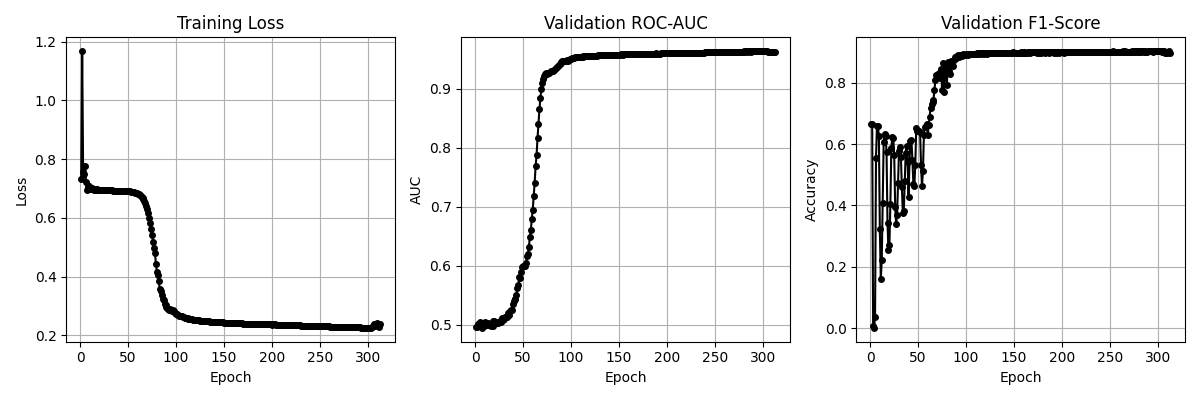}

    \caption{Training loss, validation ROC-AUC, and $F_1$-Score vs. epochs for the optimal GNN configuration on the largest cluster. Validation metrics stabilize after 100 epochs, indicating strong graph comprehension and prediction performance with increased training iterations.}
    \label{fig:opt_plots}
\end{figure*}

The optimal model configuration was selected to use LR = 0.001, weight decay = 0, hidden dimensions = 128, GNN layers = 3, and dropout = 0.1. Model training was performed for each partitioned cluster dataset. Figure~\ref{fig:opt_plots} plots the training loss, ROC-AUC, and $F_1$-Score against the number of epochs for the largest cluster. The loss plot demonstrates significant improvements with weight updates at the beginning of training, where feature refinements from random initial values are occurring. After this, the loss plateaus for approximately $50$ epochs, likely representing the model's initial understanding of the 1-hop neighbor relationships, but minimal understanding of the bipartite structure. With additional training, feature refinements eventually led to learning of the 2-hop neighbor and structural relationships, where a second drop in loss reaches the true convergence minimum.

The validation ROC-AUC plot demonstrates a maximization of prediction ranking with additional training. Initial ranking for the first $50$ epochs is poor, aligning with the stagnation of the model's understanding of complex relationships. When the model achieves greater graph understanding between $50$ and $100$ epochs, significant improvements in ranking ability occurs in the ROC-AUC plot. Stabilization and convergence occurs at a strong ROC-AUC value of $0.965$. This trend is similarly exhibited in the $F_1$-Score validation plot, where significant instability occurs for $50$ epochs, but begins to stabilize between $50$ and $100$ epochs. Again, convergence at a high score of $0.901$ occurs after $100$ epochs. These plots indicate that the model is capable of identifying complex graph relationships that lead to strong ROC-AUC ranking performance, and is capable of producing an ordered list of novel link probabilities. $F_1$-Score maximization suggests that positive predictions are often correct, with good coverage of identified novel links.

Table~\ref{tab:clustering_gnn_comparison} presents a performance comparison to related cluster and GNN studies. Silhouette results in deep learning pipelines are not commonly reported, utilizing laboratory or pharmacologic analyses for validation. Among the limited scores reported, our results demonstrated superior cluster quality, which was necessary for successful link prediction based on the breadth of drug classes included. 

Strong performance is similarly demonstrated for the GNN. ROC-AUC performance in \cite{Doshi2022} demonstrated performance of $0.855$ compared to $0.960$ for our broader study. The prediction accuracy of $0.9406$ from \cite{Xiao:23} is comparable to this study's $0.901$. These findings demonstrate resilience of the present approach, regardless of the fact that prediction accuracy was not prioritized. Many GNN link prediction analyses reported successful creation of a ranked probability list, which was similarly accomplished here.

Table~\ref{tab:cluster3_top} presents a sample of ranked model outputs, representing the highest-probability drug-disease links. Probability values indicate the confidence that an association exists, and a $99\%$ probability threshold was selected to minimize false positives, maximize the likelihood of finding unestablished drug purposes, and minimize dead ends. A list of all cluster-specific links exceeding $99\%$ probability is presented in Appendix B.

\begin{table}[htbp]
\centering
\tiny
\caption{Clustering and GNN performance metrics.}
\label{tab:clustering_gnn_comparison}
\renewcommand{\arraystretch}{1.2}
\setlength{\tabcolsep}{8pt}
\begin{tabular}{|p{1.7cm}|p{1.5cm}|p{1.5cm}|p{2cm}|}
\hline
\textbf{Study} & \textbf{Method} & \textbf{Drug Domain} & \textbf{Metric} \\
\hline
\multicolumn{4}{|c|}{\textbf{Clustering Results}} \\
\hline
Guo et al. (2020) & Hierarchical & Chinese Medicine & None; lab confirmed \\
Hameed et al. (2018) & Graph & Cancer & None; manual selection \\
Bansal et al. (2023)\textsuperscript{a} & Graph \& Spectral & COVID-19 & None; lab confirmed \\
Kim (2021) & NMF & Cancer & Silhouette = 0.36 \\
Wan et al. (2024) & Spectral & Cancer & Silhouette = 0.58 \\
\textbf{Current study} & \textbf{DEC} & \textbf{Multiple} & \textbf{Silhouette = 0.8850} \\
\hline
\multicolumn{4}{|c|}{\textbf{GNN Results}} \\
\hline
Doshi et al. (2022)\textsuperscript{a} & GNN (GDRnet) & COVID-19 & ROC-AUC = 0.855 \\
Xiao et al. (2023) & MPK-GNN & Cancer & Accuracy = 0.9406 \\
\textbf{Current study}\textsuperscript{a} & \textbf{Bipartite GNN} & \textbf{Multiple} & \textbf{ROC-AUC = 0.960}; \textbf{Accuracy = 0.901} \\
\hline
\end{tabular}

{\footnotesize$^{\mathrm{a}}$A list of link rankings was produced for qualitative evaluation.}
\end{table}

\begin{table}[htbp]
\centering
\tiny
\renewcommand{\arraystretch}{1.1}
\caption{Sample of confident drug–disease link predictions from the DEC-GNN pipeline with corroborating scientific evidence.}
\label{tab:cluster3_top}
\setlength{\tabcolsep}{6pt}
\begin{tabular}{|p{1.5cm}|p{2.2cm}|c|p{1.5cm}|p{0.8cm}|}
\hline
\textbf{Drug Name} & \textbf{Disease Name} & \textbf{Prob.} & \textbf{Study Type} & \textbf{Reference} \\
\hline
Amantadine & Breast Neoplasms & 0.9993 & Cellular & \cite{Maksymiuk2018} \\
FR900359 & Bronchial Hyperreactivity & 0.9977 & Animal & \cite{matthey2017targeted} \\
PFOSA & Angina, Stable & 0.9972 & Serologic & \cite{huang2018serum} \\
abrine & Angioedema & 0.9970 & No evidence & --\\
Cuprizone & Sarcoma, Kaposi & 0.9970 & No evidence & --\\
PFOSA & Gastritis, Atrophic & 0.9968 & No evidence & --\\
chlorantranilipole & Disease Models, Animal & 0.9967 & Animal & \cite{kimura2023neurotoxicity} \\
Cantharidin & Brain Neoplasms & 0.9963 & Cellular & \cite{wang2022inhibition} \\
dicyclohexyl phthal. & Graves Disease & 0.9960 & No evidence & --\\
FR900359 & Shock, Septic & 0.9957 & No evidence & -- \\
Methyltestosterone & Dermatomyositis & 0.9956 & No evidence & -- \\
GGTI 2147 & Heart Failure & 0.9956 & No evidence & -- \\
lacidipine & Seizures & 0.9954 & Animal & \cite{pushpa2022anticonvulsant} \\
\hline
\end{tabular}
\end{table}

The list of potential drug purposes is promising, with a high diversity of drug and disease types. Drug instances range from marketed, natural, and homeopathic subclasses. Likewise, diseases range from cancer, cardiac, and neurologic disorders, among others. Such variance improves the novelty of these findings and may inspire further pharmacologic studies.

A pharmacologic literature analysis was conducted on the links presented in Table~\ref{tab:cluster3_top} to confirm clinical validity. External literature sources that confirm the drug-disease link from any type of clinical or laboratory study are provided. Existence of prior studies do not preclude the value of these results, as ML confirmation can improve the confidence of such findings.

\section{Conclusion}
\label{sec:conclusion}
We introduced a novel DEC and GNN-based pipeline capable of identifying the probability of drug-disease links. Clustering via DEC successfully produced a reduced-dimension embedding of multi-omic properties from a broad range of drugs. DEC hyperparameter optimization resulted in a silhouette score of $0.8550$, which was superior to related works. Visual analysis demonstrated meaningful clustering of $9,022$ drugs into $35$ clusters useful for GNN processing. A GNN model capable of producing lists of drug-disease link probabilities was developed. This model exhibited strong statistical performance, with an ROC-AUC score of $0.965$ and $F_1$-Score of $0.901$, improving confidence in outputted probabilities. Lists of link probabilities were generated, with $477$ unique drug-disease links exceeding $99\%$ confidence. This study advances drug repurposing research by exploring ML pipelines capable of identifying similarities from previously unrelated drugs, and expands the library of potential links to inspire future clinical study.

Future work could be dedicated to prediction explainability. Methods that identify which features were dominant in making link predictions could further elucidate the understanding of repurposed drugs. Additionally, drugs with similar dominant features could be compared to identify the magnitude of potential disease activity. Similarly, this study could be expanded into a toxicologic analysis since the GNN unexpectedly identified links where the disease association is actually a side effect of the drug through their bipartite associations. Diseases that are treatable by certain drugs have similar pathways represented in their features, unintentionally linking drugs to side effects presented as diseases in the original dataset.

\small
\nocite{*}
\bibliographystyle{unsrt}
\bibliography{ref}

\begin{thebibliography}{10}

\bibitem{Baig:16}
Mohammad~Hassan Baig, Khurshid Ahmad, Sudeep Roy, Jalaluddin~Mohammad Ashraf1, Mohd Adil, Mohammad~Haris Siddiqui, Saif Khan, Mohammad~Amjad Kamal, Ivo Provazník, and Inho Choi.
\newblock Computer aided drug design: Success and limitations.
\newblock {\em Current pharmaceutical design}, 22(5):572--581, 2016.

\bibitem{Schlander:21}
Michael Schlander et~al.
\newblock How much does it cost to research and develop a new drug? a systematic review and assessment.
\newblock {\em PharmacoEconomics}, 39(11):1243--1269, 2021.

\bibitem{waring2015analysis}
Michael~J Waring, John Arrowsmith, Andrew~R Leach, Paul~D Leeson, Simon Mandrell, Robert~M Owen, Garry Pairaudeau, William~D Pennie, Stephen~D Pickett, Jianmin Wang, and Oliver Wallace.
\newblock An analysis of the attrition of drug candidates from four major pharmaceutical companies.
\newblock {\em Nature Reviews Drug Discovery}, 14(7):475--486, 2015.

\bibitem{Vamathevan:19}
Jessica Vamathevan, Dominic Clark, Paul Czodrowski, Ian Dunham, Edgardo Ferran, George Lee, Bin Li, Anant Madabhushi, Parantu Shah, Michaela Spitzer, and Shanrong Zhao.
\newblock Applications of machine learning in drug discovery and development.
\newblock {\em Nature reviews Drug discovery}, 18(6):463--477, April 11 2019.

\bibitem{Yang:22}
Fan Yang, Qi~Zhang, Xiaokang Ji, Yanchun Zhang, Wentao Li, Shaoliang Peng, and Fuzhong Xue.
\newblock Machine learning applications in drug repurposing.
\newblock {\em Interdisciplinary Sciences: Computational Life Sciences}, 14(1):15--21, 2022.

\bibitem{ghandikota2024application}
Srinivas~K. Ghandikota and Ashok~G. Jegga.
\newblock Application of artificial intelligence and machine learning in drug repurposing.
\newblock {\em Progress in Molecular Biology and Translational Science}, 205:171--211, 2024.

\bibitem{alshami2024smart}
Ali~K AlShami, Ryan Rabinowitz, Khang Lam, Yousra Shleibik, Melkamu Mersha, Terrance Boult, and Jugal Kalita.
\newblock Smart-vision: survey of modern action recognition techniques in vision.
\newblock {\em Multimedia Tools and Applications}, pages 1--72, 2024.

\bibitem{Tanoli:21}
Ziaurrehman Tanoli, Markus V{\"a}h{\"a}-Koskela, and Tero Aittokallio.
\newblock Artificial intelligence, machine learning, and drug repurposing in cancer.
\newblock {\em Expert Opinion on Drug Discovery}, 16(9):977--989, 2021.

\bibitem{Mohammadzadeh-Vardin2024}
Taha Mohammadzadeh-Vardin, Amin Ghareyazi, Ali Gharizadeh, Karim Abbasi, and Hamid~R. Rabiee.
\newblock Deepdra: Drug repurposing using multi-omics data integration with autoencoders.
\newblock {\em PLoS One}, 19(7):e0307649, Jul 26 2024.

\bibitem{10.1093/bib/bbab048}
Yongcui Wang, Yingxi Yang, Shilong Chen, and Jiguang Wang.
\newblock Deepdrk: a deep learning framework for drug repurposing through kernel-based multi-omics integration.
\newblock {\em Briefings in Bioinformatics}, 22(5):bbab048, 04 2021.

\bibitem{9669314}
Ruiwei Feng, Yufeng Xie, Minshan Lai, Danny~Z. Chen, Ji~Cao, and Jian Wu.
\newblock Agmi: Attention-guided multi-omics integration for drug response prediction with graph neural networks.
\newblock In {\em 2021 IEEE International Conference on Bioinformatics and Biomedicine (BIBM)}, pages 1295--1298, 2021.

\bibitem{Hameed2018}
Pathima~Nusrath Hameed, Karin Verspoor, Snezana Kusljic, and Saman Halgamuge.
\newblock A two-tiered unsupervised clustering approach for drug repositioning through heterogeneous data integration.
\newblock {\em BMC Bioinformatics}, 19(1):129, Apr 11 2018.

\bibitem{Guo:20}
Feifei Guo, Xuan Tang, Wen Zhang, Junying Wei, Shihuan Tang, Hongwei Wu, and Hongjun Yang.
\newblock Exploration of the mechanism of traditional chinese medicine by ai approach using unsupervised machine learning for cellular functional similarity of compounds in heterogeneous networks, xiaoerfupi granules as an example.
\newblock {\em Pharmacological Research}, 160:105077, 2020.

\bibitem{Doshi2022}
Siddhant Doshi and Sundeep~Prabhakar Chepuri.
\newblock A computational approach to drug repurposing using graph neural networks.
\newblock {\em Computers in Biology and Medicine}, 150:105992, 2022.

\bibitem{bansal2023clusteringgraphdeeplearningbased}
Chaarvi Bansal, Rohitash Chandra, Vinti Agarwal, and P.~R. Deepa.
\newblock A clustering and graph deep learning-based framework for covid-19 drug repurposing, 2023.

\bibitem{Xiao:23}
Shunxin Xiao, Huibin Lin, Conghao Wang, Shiping Wang, and Jagath~Chandana Rajapakse.
\newblock Graph neural networks with multiple prior knowledge for multi‑omics data analysis.
\newblock {\em IEEE Journal of Biomedical and Health Informatics}, 27(9):4591--4600, 2023.

\bibitem{ctd}
Allan~P. Davis et~al.
\newblock The comparative toxicogenomics database: update 2021.
\newblock {\em Nucleic Acids Research}, 49(D1):D1138--D1143, 2021.

\bibitem{pubchem}
Sunghwan Kim et~al.
\newblock Pubchem in 2021: new data content and improved web interfaces.
\newblock {\em Nucleic Acids Research}, 49(D1):D1388--D1395, 2021.

\bibitem{biosnap}
Stanford Network~Analysis Project.
\newblock Biosnap: A dataset for biomedical network analysis.
\newblock {\em Bioinformatics Journal}, 37(4):512--519, 2021.

\bibitem{openfda}
FDA Open~Data Initiative.
\newblock Openfda: An open access platform for drug safety.
\newblock {\em U.S. Food and Drug Administration Database}, 2020.

\bibitem{hmdb}
David~S. Wishart et~al.
\newblock Hmdb 5.0: the human metabolome database for 2022.
\newblock {\em Nucleic Acids Research}, 50(D1):D622--D631, 2022.

\bibitem{lee2020biobert}
Jinhyuk Lee, Wonjin Yoon, Sungdong Kim, Donghyeon Kim, Sunkyu Kim, Chan~Ho So, and Jaewoo Kang.
\newblock Biobert: a pre-trained biomedical language representation model for biomedical text mining.
\newblock {\em Bioinformatics}, 36(4):1234--1240, 2020.

\bibitem{wang2019smiles}
Lingling Wang, Ren Wang, Peng Guo, Xiaohong Wang, and Zhihua Wang.
\newblock Smiles-bert: Large scale unsupervised pre-training for molecular property prediction.
\newblock {\em arXiv preprint arXiv:1907.12461}, 2019.

\bibitem{kingma2015adam}
Diederik~P Kingma and Jimmy Ba.
\newblock Adam: A method for stochastic optimization.
\newblock {\em International Conference on Learning Representations (ICLR)}, 2015.

\bibitem{fey2019fastgraphrepresentationlearning}
Matthias Fey and Jan~Eric Lenssen.
\newblock Fast graph representation learning with pytorch geometric, 2019.

\bibitem{hamilton2018inductiverepresentationlearninglarge}
William~L. Hamilton, Rex Ying, and Jure Leskovec.
\newblock Inductive representation learning on large graphs, 2018.

\bibitem{Maksymiuk2018}
Andrew~W Maksymiuk, Seema~C Narayanapillai, Diane Hammond, Ramakrishna Bhat, and Krishnan Merchant.
\newblock Use of amantadine as substrate for ssat-1 activity as a reliable clinical diagnostic assay for breast and lung cancer.
\newblock {\em Future Science OA}, 5(2):FSO365, dec 2018.

\bibitem{matthey2017targeted}
Michaela Matthey, Nicholas Roberts, Chun~Yuen Seow, Jason~K Rainey, Cornelis Van~Breemen, Peter Chidiac, Boris Reidel, Susanne Dipp, Susanne Lutz, and Nils Wettschureck.
\newblock Targeted inhibition of gq signaling induces airway relaxation in mouse models of asthma.
\newblock {\em Science Translational Medicine}, 9(407):eaag2288, 2017.

\bibitem{huang2018serum}
Mengmeng Huang, Ling Wang, Haiyan Hu, Xiaomin Luo, Weihe Li, Jianjun Huang, Renjie Chen, Shunqing Xu, and Xian Wang.
\newblock Serum polyfluoroalkyl chemicals are associated with risk of cardiovascular diseases in national us population.
\newblock {\em Environment International}, 119:37--46, 2018.

\bibitem{kimura2023neurotoxicity}
Mako Kimura, Rie Kudo, Ayaka Yamashita, Kaoru Okamoto, Rie Muraoka, Satomi Matsumoto, Noriyuki Nagano, and Hidetaka Ito.
\newblock Neurotoxicity and behavioral disorders induced in mice by acute exposure to the diamide insecticide chlorantraniliprole.
\newblock {\em The Journal of Veterinary Medical Science}, 85(4):497--506, 2023.

\bibitem{wang2022inhibition}
Chie-Hong Wang, Chien-Lun Lin, Heng-Cheng Hsu, Chun-Hao Tai, Jui-Lung Chen, and Ming-Hong Lin.
\newblock Inhibition of mzf1/c-myc axis by cantharidin impairs cell proliferation in glioblastoma.
\newblock {\em International Journal of Molecular Sciences}, 23(23):14727, 2022.

\bibitem{pushpa2022anticonvulsant}
V.~H. Pushpa, M.~K. Jayanthi, S.~Mallikarjun, and Ramith Ramu.
\newblock Anticonvulsant activity of lacidipine against mes- and ptz-induced seizures: Pharmaceutical science-pharmacology for drug discovery.
\newblock {\em International Journal of Life Science and Pharma Research}, 10(5):71--75, June 2022.

\bibitem{Born:27}
Max Born and J.~Robert Oppenheimer.
\newblock On the quantum theory of molecules.
\newblock {\em Annalen der Physik}, 84(4):457--484, 1927.

\bibitem{Chen:16}
Xing Chen, Chenggang~Clarence Yan, Xiaotian Zhang, Xu~Zhang, Feng Dai, Jian Yin, and Yongdong Zhang.
\newblock Drug–target interaction prediction: databases, web servers and computational models.
\newblock {\em Briefings in bioinformatics}, 17(4):696--712, 2016.

\bibitem{Chen:19}
Chi Chen, Weike Ye, Yunxing Zuo, Chen Zheng, and Shyue~Ping Ong.
\newblock Graph networks as a universal machine learning framework for molecules and crystals.
\newblock {\em Chemistry of Materials}, 31(9):3564--3572, 2019.

\bibitem{Dara:21}
S.~Dara, S.~Dhamercherla, S.~S. Jadav, C.~M. Babu, and M.~J. Ahsan.
\newblock Machine learning in drug discovery: A review.
\newblock {\em Artificial Intelligence Review}, 55(3):1947--1999, August 11 2021.

\bibitem{Gori:05}
Marco Gori, Gabriele Monfardini, and Franco Scarselli.
\newblock A new model for learning in graph domains.
\newblock In {\em Proceedings 2005 IEEE international joint conference on neural networks}, volume~2, pages 729--734. IEEE, 2005.

\bibitem{Jain:13}
Anubhav Jain, Shyue~Ping Ong, Geoffroy Hautier, Wei Chen, William~Davidson Richards, Stephen Dacek, Shreyas Cholia, Dan Gunter, David Skinner, Gerbrand Ceder, and Kristin~A. Persson.
\newblock Commentary: The materials project: A materials genome approach to accelerating materials innovation.
\newblock {\em APL Materials}, 1(1), 2013.

\bibitem{Shermukhamedov:23}
Shokirbek Shermukhamedov, Dilorom Mamurjonova, and Michael Probst.
\newblock Structure to property: Chemical element embeddings and a deep learning approach for accurate prediction of chemical properties.
\newblock {\em arXiv preprint arXiv:2309.09355}, September 17 2023.

\bibitem{Xie:16}
Junyuan Xie, Ross Girshick, and Ali Farhadi.
\newblock Unsupervised deep embedding for clustering analysis.
\newblock In {\em International conference on machine learning}, pages 478--487, April 11 2016.

\bibitem{krishnamurthy2022drug}
Nithya Krishnamurthy et~al.
\newblock Drug repurposing: a systematic review on root causes, barriers and facilitators.
\newblock {\em BMC Health Services Research}, 22(1):970, July 2022.

\bibitem{drugbank}
David~S. Wishart et~al.
\newblock Drugbank 2022: the comprehensive, annotated, open data drug database.
\newblock {\em Nucleic Acids Research}, 50(D1):D621--D629, 2022.

\bibitem{xie2016dec}
Junyuan Xie, Ross Girshick, and Ali Farhadi.
\newblock Unsupervised deep embedding for clustering analysis.
\newblock In {\em Proceedings of the 33rd International Conference on Machine Learning}, pages 478--487, 2016.

\bibitem{kim}
Shinuk Kim.
\newblock Inferring drug set and identifying the mechanism of drugs for pc3.
\newblock {\em International Journal of Molecular Sciences}, 25:765, 01 2024.

\bibitem{Wan2024}
Zhiyu Wan, Nan Jiang, Mengqi Su, Xiaoyan Zhang, Yujie Cao, Ang Wu, Peng Zhang, and Tao Jiang.
\newblock Multiscale fusion network drives the repurposing of anticancer drugs.
\newblock {\em Clinical and Translational Medicine}, 14(7):e1745, July 2024.

\end{thebibliography}

\end{document}